\documentclass[conference]{IEEEtran}
\IEEEoverridecommandlockouts
\usepackage{cite}
\usepackage{amsmath,amssymb,amsfonts}
\usepackage{algorithmic}
\usepackage{graphicx}
\usepackage{textcomp}
\usepackage{booktabs}
\usepackage{xcolor}
\usepackage{graphicx}
\usepackage{float}
\usepackage{color}
\usepackage[most]{tcolorbox}
\def\BibTeX{{\rm B\kern-.05em{\sc i\kern-.025em b}\kern-.08em
    T\kern-.1667em\lower.7ex\hbox{E}\kern-.125emX}}
\begin{document}

\title{Exploring the In-Context Learning Capabilities of LLMs for Money Laundering Detection in Financial Graphs
\thanks{\textcopyright 2025 The Vanguard Group, Inc. All rights reserved.}
}

\author{\IEEEauthorblockN{Erfan Pirmorad}
\IEEEauthorblockA{\textit{The Vanguard Group, Inc.} \\
erfan\_pirmorad@vanguard.com}}


\maketitle

\begin{abstract}
The complexity and inter-connectivity of entities involved in money laundering demand investigative reasoning over graph-structured data. This paper explores the use of large language models (LLMs) as reasoning engines over localized subgraphs extracted from a financial knowledge graph. We propose a lightweight pipeline that retrieves k-hop neighborhoods around entities of interest, serializes them into structured text, and prompts an LLM via few-shot in-context learning to assess suspiciousness and generate justifications. Using synthetic anti-money laundering (AML) scenarios that reflect common laundering behaviors, we show that LLMs can emulate analyst-style logic, highlight red flags, and provide coherent explanations. While this study is exploratory, it illustrates the potential of LLM-based graph reasoning in AML and lays groundwork for explainable, language-driven financial crime analytics.
\end{abstract}



\section{Introduction}
As financial data becomes increasingly interconnected, institutions are turning to knowledge graphs to capture the complex relationships between customers, accounts, transactions, devices, and geographic entities. These graphs provide a rich structure for reasoning about risk, behavior, and anomalies. However, leveraging their full potential requires tools capable of understanding and explaining multi-hop, heterogeneous patterns. 

Recent advances in large language models (LLMs) have demonstrated their ability to reason over structured data, including relational chains and graph-like abstractions. A growing body of research has shown that LLMs can be adapted to perform graph reasoning tasks via graph-structured prompting \cite{Fatemi2023TalkLA}, or language-guided message passing \cite{Perozzi2024LetYG}. Surveys on this topic highlight a promising direction for unifying natural language and graph reasoning under a shared framework \cite{Pan2023UnifyingLL}.

Building on this momentum, we explore whether LLMs can support investigative-style reasoning over \textit{financial knowledge graphs}. These graphs represent real-world financial systems, where relational complexity is not only common but essential to risk analysis, trend analysis, etc., \cite{Li2024FinDKGDK}. Our goal is to evaluate the feasibility of using LLMs as interpretable reasoning engines over such data. As a motivating case study, we focus on anti-money laundering (AML) in this study. AML is a domain where identifying suspicious behavior involves tracing intricate financial flows across individuals, shell entities, and intermediary accounts \cite{10.1007/s10115-017-1144-z, DOMASHOVA2021184}. AML and financial crime investigations increasingly rely on graph-based representations to detect patterns such as layering, rapid movement of funds, and shared infrastructure \cite{OZTAS2024161, MOTIE2024122156}. 

For this study, we utilize the publicly available IBM AML Synthetic Dataset~\cite{altman2023realistic}, which simulates financial activities through a realistic knowledge graph of accounts, banks, and transactions, designed to model common laundering typologies based on real-world intelligence. This allows us to explore LLM-based reasoning in a controlled yet representative environment.

This work positions LLMs not merely as text generators, but as \textit{in-context reasoning engines} capable of interpreting graph-structured financial activity. In-context learning offers a flexible, parameter-free mechanism to inject domain knowledge and reasoning capabilities into black-box LLMs, aligning naturally with investigative workflows. By leveraging this mechanism, we evaluate whether LLMs can emulate analyst-style reasoning in AML detection without retraining or labeled supervision.

Our pipeline (Fig. 1) extracts a k-hop subgraph, serializes it into structured text, and prompts an LLM in a few-shot setup to classify suspiciousness and provide human-readable justification.

\begin{figure}[H]
\centering
\includegraphics[width=\linewidth]{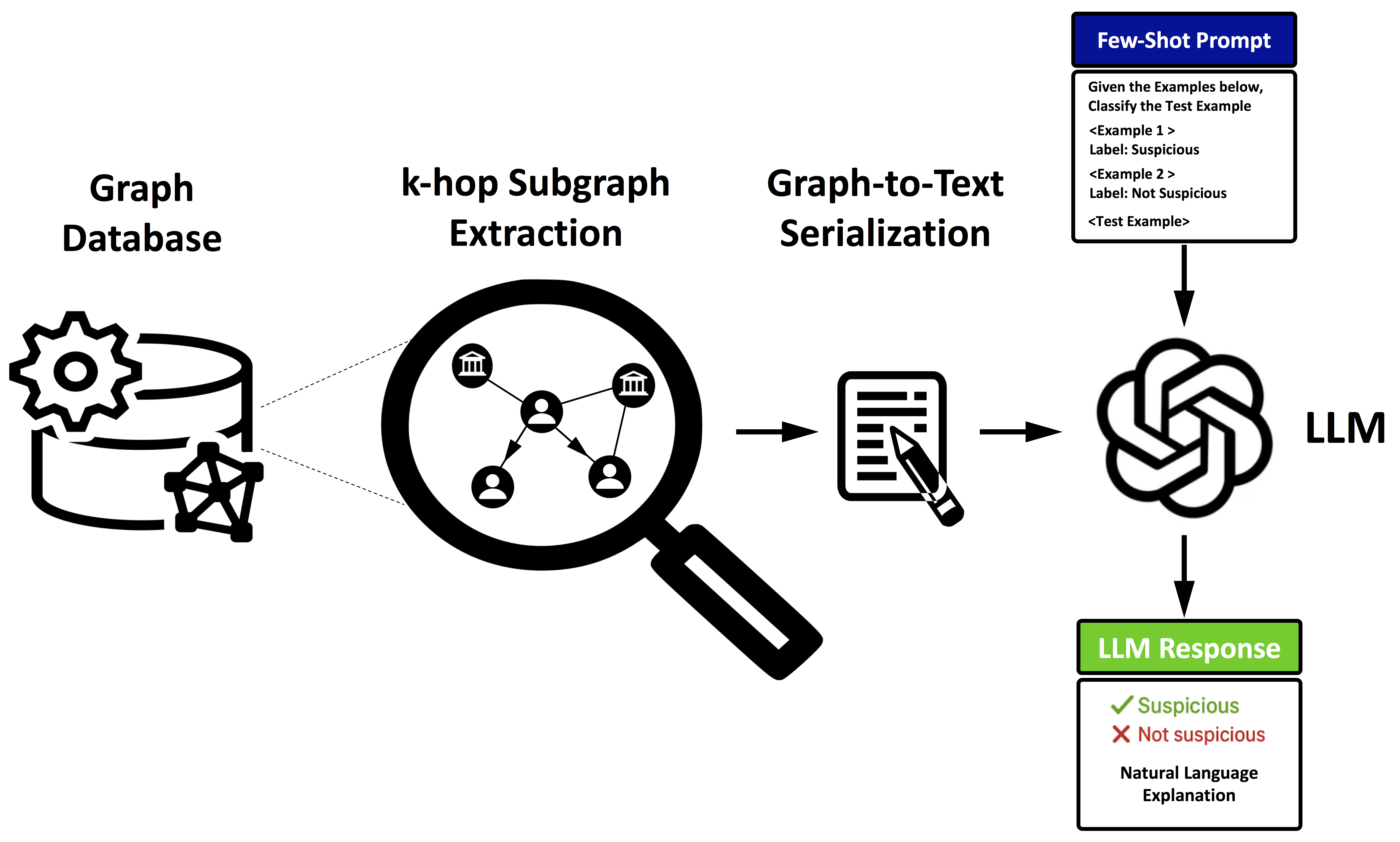}
\caption{Pipeline Overview for LLM-based AML Analysis.}
\label{fig:pipeline-architecture}
\end{figure}

We evaluate this pipeline using synthetic AML scenarios that reflect realistic laundering behaviors, constructed from publicly available datasets. Our early results indicate that LLMs can highlight meaningful red flags, trace multi-step patterns, and generate rationales that align with human intuition. This case study illustrates the potential of LLMs for reasoning over financial knowledge graphs, offering a path toward explainable, language-driven fraud analytics. To our knowledge, this is the first work to apply in-context learning with LLMs for money laundering detection over financial graphs. The proposed pipeline bridges localized graph structure with natural language reasoning in a scalable and interpretable way.

\section{Methodology}

We propose a three-stage framework: (1) extract subgraphs around suspicious transactions, (2) serialize them into text, and (3) prompt an LLM with laundering typologies in a few-shot setup. The model outputs suspiciousness, rationale, and typology, letting us evaluate classification accuracy and reasoning transparency.

\subsection{Financial Graph Representation and Subgraph Extraction}

Money laundering schemes often involve complex transactional relationships that can be naturally represented as graphs. In this work, we use the publicly available IBM AML Synthetic Dataset~\cite{altman2023realistic}, which models financial activities as a knowledge graph. In this graph, nodes represent \textit{accounts} (individual or corporate) and \textit{banks}, while edges represent \textit{financial transactions} between accounts. Each node is annotated with metadata such as entity type and account creation date; each transaction edge is labeled with attributes including transaction amount and timestamp.

Our goal is to reason about the suspiciousness of specific transactions by analyzing the surrounding graph structure. For each candidate transaction (i.e., a transfer between two accounts), we extract a localized $k$-hop subgraph centered on the transaction edge. The subgraph includes the source and destination accounts, their neighboring transactions within $k$ hops, and any associated bank nodes. An example of an extracted subgraph is shown below. Each extracted structure provides the LLM with a focused local context to assess whether the transaction may reflect suspicious activity.

\begin{figure}[h]
\centering
\includegraphics[width=5cm]{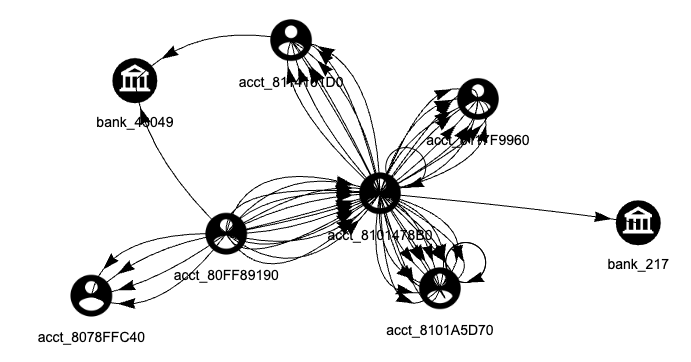}
\caption{Snapshot of a sample 2-hop subgraph centered around a transaction between \textit{acct\_80FF89190} and \textit{acct\_810147BB0}.}
\label{fig:pipeline-architecture}
\end{figure}

This extraction strategy frames the detection task as an \textit{edge-centric classification problem} over the transaction graph, where the objective is to predict whether the given transaction edge represents money laundering activity.

\subsection{Textual Serialization}\label{subsec:text-serialization}

Each subgraph is serialized into structured text: nodes as tagged entities and edges as relations with direction, amounts, and timestamps. Example:

\begin{tcolorbox}[colback=gray!5, colframe=gray!80!black,
  title=Serialized Subgraph Example,
  fontupper=\scriptsize,
  boxsep=2pt, left=2pt, right=2pt, top=2pt, bottom=2pt,
  before upper={\linespread{0.85}\selectfont}
]
\ttfamily
\textbf{\texttt{**Nodes:**}}\par
- acct\_810147BB0 (type: Account)\par
- acct\_8101A5D70 (type: Account)\par
- acct\_811416D0 (type: Account)\par
- acct\_8117F9960 (type: Account)\par
- acct\_80FF8910 (type: Account)\par
- bank\_217 (type: Bank)\par
- bank\_4049 (type: Bank)\par

\vspace{0.1em}
\textbf{\texttt{**Edges:**}}\par
- acct\_810147BB0 belongs\_to bank\_217\par

\vspace{0.1em}
- acct\_810147BB0 transfers\_to acct\_8101A5D70\par
\hspace*{1em}amount: 225756.22 Shekel\par
\hspace*{1em}via: Reinvestment\par
\hspace*{1em}timestamp: 2022/09/01 00:02\par

\textit{...additional edges omitted for brevity...}
\end{tcolorbox}

\subsection{Few-Shot Prompting with AML Typologies}\label{subsec:few-shot-prompting}

\begin{figure}[H]
\centering
\includegraphics[width=\linewidth]{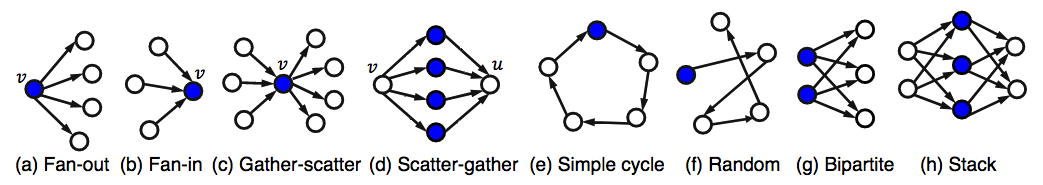}
\caption{Canonical money laundering patterns used as reasoning demonstrations. Image adapted from the IBM AML dataset paper.}
\label{fig:aml_patterns}
\end{figure}

A defining characteristic of anti-money laundering (AML) investigations is the presence of recurring structural patterns in transactional behavior. Some of these patterns have been studied and documented in financial intelligence reports and academic studies, including the IBM AML dataset, which models them as prototypical graph topologies. Figure~\ref{fig:aml_patterns} visualizes eight canonical laundering structures used in our prompting framework: (a) \textit{fan-out}, (b) \textit{fan-in}, (c) \textit{gather-scatter}, (d) \textit{scatter-gather}, (e) \textit{simple cycle}, (f) \textit{random}, (g) \textit{bipartite}, and (h) \textit{stack}.

To help the LLM learn to reason over financial subgraphs, our approach uses a few-shot prompting setup where each demonstration example showcases one of these laundering patterns. Each example consists of (i) a brief serialized subgraph in the textual format described in Section~\ref{subsec:text-serialization}, and (ii) a natural language rationale that explains why the activity is suspicious, grounded in the structure of the graph. Both the graph serialization and the rationale are derived from samples in the IBM AML dataset, which provides well-annotated synthetic cases.

The prompt includes eight such examples, each corresponding to one of the patterns shown in Figure~\ref{fig:aml_patterns}, before appending the test case. By explicitly conditioning the LLM on real-world-style typologies and human-readable rationales, we align the reasoning process with investigative workflows. Importantly, these demonstrations emphasize graph topology and multi-hop interactions rather than superficial statistical cues, which makes the model more likely to generalize beyond seen entities.

This methodology leverages the core advantage of few-shot learning via in-context learning: enabling the model to acquire structural priors without any fine-tuning. It effectively bootstraps the LLM into a role akin to a fraud analyst by showing what suspicious behavior “looks like” in multiple formats. The choice to use real pattern types, realistic transactions, and natural language justifications makes the prompting setup robust, interpretable, and well-aligned with human investigator expectations. To further refine the model’s discrimination ability, we also include several examples of non-suspicious financial activity—subgraphs that contain no red flags—within the same prompt. This ensures that the model learns not only to detect suspicious patterns, but also to distinguish them from legitimate transactional behavior.

The final structure of the LLM prompt follows the template below, showing how demonstration subgraphs and rationales are interleaved. The subgraphs are serialized using the format shown previously in the \textit{Serialized Subgraph Example} box in Section~\ref{subsec:text-serialization}.

\begin{tcolorbox}[colback=white, colframe=blue!60!black,
  title=Final Prompt Template,
  fontupper=\scriptsize,
  boxsep=2pt, left=2pt, right=2pt, top=2pt, bottom=2pt,
  before upper={\linespread{0.85}\selectfont}
]
 
\ttfamily
\textbf{Prompt Header:} \par
You are an expert financial crime investigator reviewing patterns of financial activities and behaviors of involved accounts to identify potential cases of money laundering. The data is represented as a graph, where: \par
\textbf{- Nodes} are of type \texttt{Account} or \texttt{Bank}. \par
\textbf{- Edges} represent relationships of type \texttt{transfers\_to} or \texttt{belongs\_to}, and include metadata such as \texttt{amount}, \texttt{currency}, \texttt{payment method}, and \texttt{timestamp}. \\ 

For training purposes, you will be shown examples of subgraph typologies that are known to be either suspicious (indicative of laundering tactics) or non-suspicious (routine financial activity). These typologies illustrate common structural patterns in financial networks.

\vspace{0.7em}
\textbf{Few-shot Examples:}

<<Serialized Subgraph 1>> \par
\textbf{Explanation:} This pattern resembles fan-out behavior, where a single account disperses funds to many accounts within a short window, often to obscure the origin.

\vspace{0.4em}
<<Serialized Subgraph 2>> \par
\textbf{Explanation:} This reflects a gather-scatter pattern, where funds from multiple sources converge briefly before dispersing again, which is a common layering tactic.

\vspace{0.8em}
... (examples 3 through 7 omitted for brevity) ...

\vspace{0.8em}
<<Serialized Subgraph 8>> \par
\textbf{Explanation:} A highly connected bipartite structure, consistent with mule networks or coordinated burst activity.

\vspace{0.7em}
\textbf{non-suspicious Examples:}

<<Serialized Subgraph 1>> \par
\textbf{Explanation:} These transfers occur across a regular schedule, between known parties with consistent values. No indicators of laundering behavior.

\vspace{0.8em}
... (examples 2 through 4 omitted for brevity) ...

\vspace{0.7em}

\textbf{Task:} Given a transaction (edge) with Transaction ID <ID>, along with its surrounding subgraph, determine whether the transaction is suspicious, or not suspicious. Your reasoning should be based on whether the surrounding subgraph resembles any of the suspicious typologies provided in the examples.\par
\textbf{Test Example:}

<<Serialized Test Subgrap>>\par

\textbf{Answer Format:}\par
- Conclusion: Suspicious or Not Suspicious \par
- Explanation: (2-3 sentences reasoning)\par
- Observed Pattern: (e.g., gather-scatter)

\end{tcolorbox}


\section{Experiments and Results}

We evaluate the proposed reasoning pipeline on the \texttt{AML HI Small} dataset, which simulates real-world money laundering typologies through graph-structured financial transactions. Our primary goal at this stage is to assess both the classification ability and the reasoning quality of a large language model (LLM) over structured financial subgraphs.

\subsection{Experimental Setup}

We construct a balanced evaluation set consisting of 2,000 examples: 1,000 laundering transactions and 1,000 non-laundering transactions. 
While this does not reflect the class imbalance typically observed in real-world AML investigations, it enables a controlled evaluation of the model’s ability to distinguish suspicious patterns without being dominated by prior class frequencies.
This design isolates the core reasoning capabilities of the LLM pipeline and ensures that performance metrics reflect meaningful classification behavior rather than dataset skew.
Extending the evaluation to realistic, highly imbalanced distributions remains an important direction for future work. Each subgraph is serialized following the format described in Section~\ref{subsec:text-serialization} and embedded into the final prompt template described in Section~\ref{subsec:few-shot-prompting}. The prompt includes eight suspicious pattern examples and four non-suspicious examples, with task instructions asking the model to: (1) classify the subgraph as suspicious or not suspicious, and (2) list any observed money laundering patterns. We use the GPT-4o model accessed via the OpenAI API (temperature = 0). Each test sample is independently prompted with no carryover memory.

\subsection{Sample Model Output}

Below we present a representative output generated by the LLM for a test subgraph classified as suspicious:
\begin{tcolorbox}[colback=green!5, colframe=green!80!black,
  title=Sample LLM Output,
  fontupper=\scriptsize,
  boxsep=2pt, left=2pt, right=2pt, top=2pt, bottom=2pt,
  before upper={\linespread{0.85}\selectfont}
]

{\ttfamily
\textbf{Conclusion:} Suspicious \par

\textbf{Explanation:} Transaction ID `3064852` is part of a networks that exhibits characteristics of fan-out. The originating account `acct\_8089DC420` rapidly transfers large amounts to multiple accounts across different banks, including the target account `acct\_807051670`. The high volume of transactions, diverse endpoints, and rapid movement suggest layering and potential laundering activity.\par

\textbf{Observed Pattern:} fan-out
}
\end{tcolorbox}

\subsection{Classification Performance}

Given the computational cost of LLM inference at scale, rerunning the model over multiple randomized test sets was infeasible. 
Instead, to assess the robustness of our results without incurring additional inference cost, we apply bootstrap resampling over the fixed set of 2,000 model outputs.
Specifically, we generate 1,000 bootstrap samples (sampling with replacement) and recompute key evaluation metrics for each resample.
This procedure allows us to estimate the mean and 95\% confidence intervals for classification accuracy, precision, and recall.
Reporting confidence intervals provides insight into the stability of both classification and reasoning performance. Table~\ref{tab:classification} summarizes the model’s binary classification performance.

\begin{table}[h]
\centering
\caption{Classification Performance with 95\% Confidence Intervals}
\label{tab:classification}
\begin{tabular}{lc}
\toprule
\textbf{Metric} & \textbf{Value (95\% CI)} \\
\midrule
Overall Accuracy & 63.7\% $\pm$ 3.3\% \\
Precision & 62.7\% $\pm$ 4.4\% \\
Recall & 67.7\% $\pm$ 4.4\% \\
\bottomrule
\end{tabular}
\end{table}

To evaluate the model’s reasoning quality and tendency to hallucinate patterns, we measure the precision and recall of the patterns listed by the LLM relative to ground-truth pattern labels provided by the IBM dataset.

\begin{table}[h]
\centering
\caption{Pattern Recognition Performance (Per Pattern) with 95\% Confidence Intervals}
\begin{tabular}{lcc}
\toprule
\textbf{Pattern} & \textbf{Precision (95\% CI)} & \textbf{Recall (95\% CI)} \\
\midrule
Fan-out & 75.1\% $\pm$ 4.0\% & 80.3\% $\pm$ 3.8\% \\
Fan-in & 73.8\% $\pm$ 4.1\% & 81.5\% $\pm$ 4.0\% \\
Gather-scatter & 66.2\% $\pm$ 4.5\% & 70.7\% $\pm$ 4.3\% \\
Scatter-gather & 64.9\% $\pm$ 4.5\% & 68.8\% $\pm$ 4.3\% \\
Simple cycle & 62.5\% $\pm$ 4.6\% & 65.9\% $\pm$ 4.5\% \\
Random & 51.5\% $\pm$ 5.0\% & 53.5\% $\pm$ 4.9\% \\
Bipartite & 55.0\% $\pm$ 5.0\% & 56.1\% $\pm$ 5.0\% \\
Stack & 52.0\% $\pm$ 5.1\% & 58.0\% $\pm$ 5.0\% \\
\bottomrule
\end{tabular}
\end{table}

Detection performance generally reflects the underlying structural complexity of the laundering patterns. Highly organized patterns such as \textit{fan-out} and \textit{fan-in} achieve the strongest results, while moderately complex patterns like \textit{gather-scatter}, \textit{scatter-gather}, and \textit{simple cycle} show intermediate performance. Among the harder patterns, such as \textit{random}, \textit{bipartite}, and \textit{stack}, we observe some variability, reflecting the increased difficulty in consistently capturing less structured behaviors. Overall, these preliminary findings suggest that prompting LLMs over serialized financial graphs provides a viable pathway for both suspicious activity classification and explainable pattern-driven reasoning.


\section{Conclusion and Future Work}

In this paper, we explored the feasibility of using large language models (LLMs) to perform structured reasoning over financial knowledge graphs for anti-money laundering (AML) investigations. By constructing localized subgraphs around transactions, serializing them into textual prompts, and conditioning the LLM with a few-shot setup based on canonical laundering patterns, we demonstrated that LLMs can not only classify suspicious behavior but also provide interpretable rationales for their decisions through in-context learning. 

While this work presents an initial exploration rather than a production-ready system, it is not intended as a fully benchmarked replacement for existing graph-based fraud detection methods. Instead, our goal was to assess the feasibility and interpretability of in-context LLM reasoning over financial knowledge graphs. This study provides a conceptual foundation for hybrid approaches that integrate explainability and investigative logic, while also recognizing that computational and scalability challenges must be addressed for future practical deployments. One of the most important advantages of this approach is its strong interpretability. Unlike traditional graph neural networks (GNNs) or other embedding-based methods, which often act as black boxes, the LLM produces human-readable explanations that align with investigative reasoning. This opens the door for models that are not only accurate but also transparent and auditable, addressing a key pain point in the deployment of machine learning systems for financial crime detection.

While promising, the approach also presents practical limitations. In particular, the computational cost of prompting large language models at inference time makes it impractical as a stand-alone solution for large-scale production. Instead, we envision a hybrid system where a lightweight classifier, such as a GNN or a graph feature engineering model with a gradient boosting classifier, performs initial triage. Only cases with low classification confidence (borderline decisions) are escalated to the LLM for additional reasoning. This selective invocation strategy maintains scalability while leveraging LLM-generated rationales to improve decision quality and explainability. In future work, we plan to benchmark such systems and assess their impact on detection performance and transparency. We will also explore whether smaller LLMs, fine-tuned for financial graph reasoning, can deliver similar interpretability with significantly reduced computational overhead. These steps will help unlock scalable, interpretable, and effective fraud and financial crime detection systems for real-world environments.


\section*{Disclaimer}

This material is provided for informational purposes only and is not intended to be investment advice or a recommendation to take any particular investment action.

\bibliographystyle{plain}
\bibliography{mybibfile}

\begin{thebibliography}{1}

\bibitem{altman2023realistic}
Erik Altman, Jovan Blanu{\v{s}}a, Luc von Niederh{\"a}usern, B{\'e}ni Egressy, Andreea Anghel, and Kubilay Atasu.
\newblock Realistic synthetic financial transactions for anti-money laundering models.
\newblock In {\em Advances in Neural Information Processing Systems 36 (NeurIPS 2023) Datasets and Benchmarks Track}, 2023.

\bibitem{10.1007/s10115-017-1144-z}
Zhiyuan Chen, Le~Dinh Khoa, Ee~Na Teoh, Amril Nazir, Ettikan~Kandasamy Karuppiah, and Kim~Sim Lam.
\newblock Machine learning techniques for anti-money laundering (aml) solutions in suspicious transaction detection: a review.
\newblock {\em Knowl. Inf. Syst.}, 57(2):245–285, November 2018.

\bibitem{DOMASHOVA2021184}
Jenny Domashova and Natalia Mikhailina.
\newblock Usage of machine learning methods for early detection of money laundering schemes.
\newblock {\em Procedia Computer Science}, 190:184--192, 2021.
\newblock 2020 Annual International Conference on Brain-Inspired Cognitive Architectures for Artificial Intelligence: Eleventh Annual Meeting of the BICA Society.

\bibitem{Fatemi2023TalkLA}
Bahare Fatemi, Jonathan~J. Halcrow, and Bryan Perozzi.
\newblock Talk like a graph: Encoding graphs for large language models.
\newblock {\em ArXiv}, abs/2310.04560, 2023.

\bibitem{Li2024FinDKGDK}
Xiaohui~Victor Li and Francesco~Sanna Passino.
\newblock Findkg: Dynamic knowledge graphs with large language models for detecting global trends in financial markets.
\newblock In {\em International Conference on AI in Finance}, 2024.

\bibitem{MOTIE2024122156}
Soroor Motie and Bijan Raahemi.
\newblock Financial fraud detection using graph neural networks: A systematic review.
\newblock {\em Expert Systems with Applications}, 240:122156, 2024.

\bibitem{OZTAS2024161}
Berkan Oztas, Deniz Cetinkaya, Festus Adedoyin, Marcin Budka, Gokhan Aksu, and Huseyin Dogan.
\newblock Transaction monitoring in anti-money laundering: A qualitative analysis and points of view from industry.
\newblock {\em Future Generation Computer Systems}, 159:161--171, 2024.

\bibitem{Pan2023UnifyingLL}
Shirui Pan, Linhao Luo, Yufei Wang, Chen Chen, Jiapu Wang, and Xindong Wu.
\newblock Unifying large language models and knowledge graphs: A roadmap.
\newblock {\em IEEE Transactions on Knowledge and Data Engineering}, 36:3580--3599, 2023.

\bibitem{Perozzi2024LetYG}
Bryan Perozzi, Bahare Fatemi, Dustin Zelle, Anton Tsitsulin, Mehran Kazemi, Rami Al-Rfou, and Jonathan~J. Halcrow.
\newblock Let your graph do the talking: Encoding structured data for llms.
\newblock {\em ArXiv}, abs/2402.05862, 2024.

\end{thebibliography}

\end{document}